\documentclass[lettersize,journal]{IEEEtran}
\usepackage{amsmath,amsfonts}
\usepackage{array}
\usepackage[caption=false,font=normalsize,labelfont=sf,textfont=sf]{subfig}
\usepackage{textcomp}
\usepackage{stfloats}
\usepackage{url}
\usepackage{verbatim}
\usepackage{graphicx}
\usepackage{cite}
\usepackage{pifont}%
\usepackage{amssymb}%
\usepackage{ulem}
\usepackage{algorithm}
\usepackage{algpseudocode}

\usepackage[table]{xcolor}
\usepackage{hyperref}
\usepackage{url}
\usepackage[utf8]{inputenc} % allow utf-8 input
\usepackage[T1]{fontenc}    % use 8-bit T1 fonts
\usepackage{xr-hyper} 
\usepackage{pdfpages}
\usepackage{selectp}
\usepackage{booktabs}       % professional-quality tables
\usepackage{amsfonts}       % blackboard math symbols
\usepackage{nicefrac}       % compact symbols for 1/2, etc.
\usepackage{microtype}      % microtypography
\usepackage{xcolor}         % colors
\usepackage{graphicx}
\usepackage{caption}
\usepackage{subcaption}
\usepackage{float}
\usepackage{wrapfig}
\usepackage{multirow}

\algrenewcommand\algorithmiccomment[1]{\textcolor{gray}{// #1}}

\begin{document}

\title{FastTracker: Real-Time and Accurate Visual Tracking}

\author{Hamidreza Hashempoor,~Yu Dong Hwang
    
        \thanks{H. Hashempoor and Y. Hwang are with with Pintel Co. Ltd., Seoul, South Korea (e-mail: \{Hamidreza, ~Ydhwang\}@pintel.co.kr). (Corresponding Author: Hamidreza Hashempoor)}}



\maketitle

\begin{abstract}
Conventional multi-object tracking (MOT) systems are predominantly designed for pedestrian tracking and often exhibit limited generalization to other object categories. This paper presents a generalized tracking framework capable of handling multiple object types, with a particular emphasis on vehicle tracking in complex traffic scenes. The proposed method incorporates two key components: (1) an occlusion-aware re-identification mechanism that enhances identity preservation for heavily occluded objects, and (2) a road-structure-aware tracklet refinement strategy that utilizes semantic scene priors—such as lane directions, crosswalks, and road boundaries—to improve trajectory continuity and accuracy. In addition, we introduce a new benchmark dataset comprising diverse vehicle classes with frame-level tracking annotations, specifically curated to support evaluation of vehicle-focused tracking methods. Extensive experimental results demonstrate that the proposed approach achieves robust performance on both the newly introduced dataset and several public benchmarks, highlighting its effectiveness in general-purpose object tracking.
While our framework is designed for generalized multi-class tracking, it also achieves strong performance on conventional benchmarks, with HOTA scores of 66.4 on MOT17 and 65.7 on MOT20 test sets. \href{https://github.com/Hamidreza-Hashempoor/FastTracker}{\textcolor{magenta}{github.com/Hamidreza-Hashempoor/FastTracker}}, \href{https://huggingface.co/datasets/Hamidreza-Hashemp/FastTracker-Benchmark}{\textcolor{magenta}{huggingface.co/datasets/Hamidreza-Hashemp/FastTracker-Benchmark}}.
\end{abstract}


\section{Introduction}
\IEEEPARstart{M}{ulti} object tracking (MOT) plays a critical role in computer vision applications such as intelligent surveillance and autonomous driving. Despite significant progress, it remains a challenging problem due to factors like target similarity, frequent occlusions, and the continuous entry and exit of objects from the scene~\cite{milan2016mot16}. A widely adopted approach to address MOT is the \textit{tracking-by-detection} paradigm~\cite{babaee2017combined}, where object detectors first identify candidate targets in each frame, and a separate tracking module associates these detections with existing trajectories. This association is typically formulated as a matching problem, relying on computed similarity measures between current detections and existing tracklets. Algorithms such as the Hungarian method~\cite{bewley2016simple}, 
 have been employed to solve this assignment task effectively. Detection algorithms also provide a confidence score for each bounding box, which reflects the likelihood of a detection being valid. In general, high-confidence detections are expected to correspond to true positives, while lower scores often indicate false positives.

\begin{figure}[t]
    \centering
    \includegraphics[width=\linewidth, keepaspectratio]{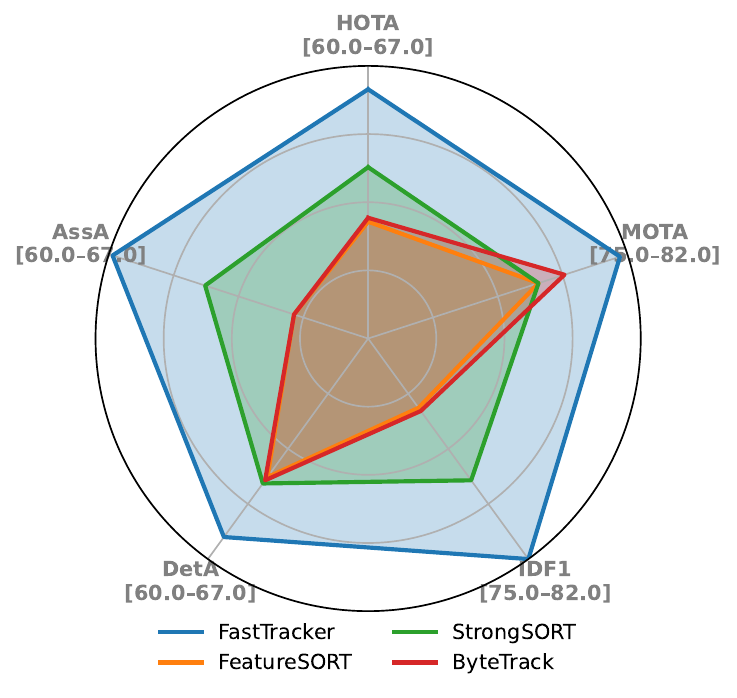}
    \caption{Comparison of multi-object tracking performance using five key metrics: HOTA, MOTA, IDF1, DetA, and AssA on MOT17 test set. Each axis is within the specified range, highlighting trade-offs between detection accuracy
and association quality.}
    \label{fig:tracker_radar}
\end{figure}

In general, high-confidence detections are expected to correspond to true positives, while lower scores often indicate false positives. However, the performance of many existing tracking-by-detection frameworks is often overrated, as they are primarily benchmarked on single-class scenarios—most notably pedestrian tracking—where detectors are trained and optimized specifically for that class \cite{babaee2017combined}. While such specialization can yield impressive accuracy, it does not reflect the challenges of real-world applications, where diverse object categories must be simultaneously detected and tracked. When detection models are extended to handle multiple object classes, detection accuracy typically decreases, leading to a corresponding drop in tracking performance \cite{hao2024divotrack}. This discrepancy highlights the need for tracking solutions that generalize well to multi-class settings and remain robust under more realistic deployment conditions.

Building on the limitations of single-class tracking, multi-class object tracking requires careful consideration of confidence distributions across different categories. In practice, high-confidence detections are generally more reliable and typically correspond to true positives, whereas low-confidence detections are prone to false positives. This disparity in reliability has motivated cascaded matching strategies, such as the one employed in ByteTrack~\cite{zhang2022bytetrack}, where high-confidence detections are prioritized in the initial matching stage, followed by selective matching of low-confidence candidates to previously unmatched trajectories.

Inspired by this approach, our tracking framework introduces a two-stage matching process that explicitly distinguishes between high- and low-confidence detections. In the first stage, we associate high-confidence detections with active tracklets using a relaxed similarity threshold, thereby maximizing the recall of true positive associations. In the second stage, we process the remaining low-confidence detections using a stricter similarity constraint, ensuring that only the most plausible associations are considered for tracks that remained unmatched.

To evaluate the similarity between detections and tracklets, our method primarily relies on motion cues, including spatial proximity, bounding box geometry, and velocity consistency across frames. While appearance-based features are commonly used in many tracking systems to enhance robustness under occlusion, they typically require deep convolutional networks that introduce considerable computational overhead \cite{yu2016poi}. This makes them less suitable for real-time, online applications. Instead of relying on such heavy models \cite{hashempoor2024featuresort}, our approach adopts two complementary strategies to improve robustness in challenging scenarios, which do not rely on CNN-based ReID networks. First, we introduce an occlusion-handling framework that enables re-identification of objects after temporary disappearance. Second, we leverage high-level environmental context—such as road layouts, two-way traffic structures, and pedestrian crosswalks—to guide re-identification and refine tracklet trajectories. These two components significantly improve tracking reliability without incurring high computational costs. The overall performance trends of our tracker compared to existing methods, measured across multiple metrics, are illustrated in Figure~\ref{fig:tracker_radar}.

\begin{figure}[t]
    \centering
    \includegraphics[width=\linewidth, keepaspectratio]{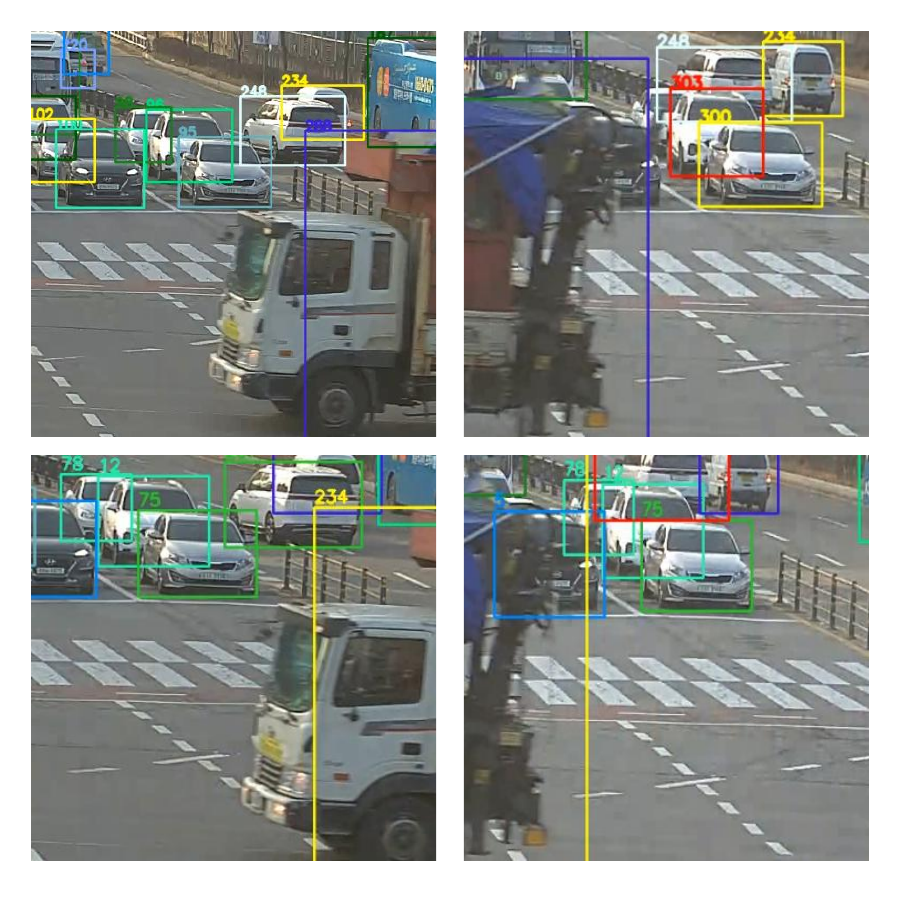}
    \caption{
    Tracking results without (top) and with (bottom) occlusion handling. Our method preserves vehicle IDs after severe occlusion, while the baseline shows frequent ID switches.
    }
    \label{fig:fasttrack_occ_demo}
\end{figure}

As the first component, we address the challenge of occlusion in multi-object tracking by designing a mechanism that operates without relying on visual re-identification features. When a target temporarily disappears from detection, we use its confidence history and spatial interactions with nearby objects to infer occlusion events. In cases where objects from the same class occlude one another, we propose a geometric overlap-based heuristic to identify covered targets. Our method introduces a novel coverage metric to more accurately detect occlusions in scenarios where traditional IoU-based measures fall short, particularly when size disparities exist between occluding and occluded objects. Once a target is marked as occluded, we adapt its Kalman filter update by moderating velocity and size changes, preventing unrealistic drift and ensuring stability during re-identification. This strategy enables more reliable identity preservation in crowded and dynamic environments. The effectiveness of our proposed occlusion handling is visually demonstrated in Figure \ref{fig:fasttrack_occ_demo}.

To further improve tracking consistency, we incorporate scene-level structural priors derived from the surrounding environment. In particular, we model roads and crosswalks as rectangular regions, where each region is annotated with dominant entry and exit edges. These edge pairs define an expected movement direction, which we use to estimate a trajectory angle threshold for objects within that region. For example, in one-way roads, the allowed motion is constrained to a narrow angular band aligned with traffic flow; significant deviation from this angle is considered physically implausible. During tracking, if a trajectory violates this contextual constraint, we project it back into the allowable motion corridor to correct potential drifts. This projection is especially useful in recovering from tracking failures caused by occlusion or detector noise, where misaligned trajectories often result from incorrect Kalman updates or false detections. By enforcing environment-aware motion consistency, we achieve more reliable tracklet associations and reduce identity switches in structured traffic scenes.

While significant progress has been made in multi-object tracking, most existing benchmarks—such as MOT17 \cite{dendorfer2020motchallenge} and MOT20 \cite{dendorfer2020mot20}—are focused almost exclusively on pedestrian tracking. Some datasets target vehicle tracking \cite{geiger2012we, sun2020scalability, tang2019cityflow}; however, they typically involve a few object classes and are captured in relatively simple driving environments, lacking the complexity of real-world traffic scenarios without CCTV view in different hours with various light conditions (including both day and night times). 
To address this gap, we introduce a new benchmark specifically designed for multi-class tracking that includes both pedestrians and various types of vehicles from urban CCTV views. Our dataset features a diverse set of object categories—such as cars, multiple truck types, buses, motorcycles, and others—alongside pedestrian targets in complex urban environments. It captures challenging scenarios with frequent occlusions, dense intersections, and multi-directional object movement, providing a realistic and demanding setting for evaluating multi-class tracking algorithms.
By capturing complex interactions across varied environments, this benchmark provides a more realistic evaluation setting and encourages the development of tracking algorithms that generalize beyond pedestrian-only settings that offers a valuable resource for the research community and facilitates progress in multi-class tracking.

The main contributions of this work are summarized as follows:

\begin{itemize}
    \item We propose a robust multi-class multi-object tracking framework that generalizes beyond pedestrian tracking and performs effectively on various vehicle classes in complex urban environments.
    
    \item We design a lightweight occlusion-handling module that does not rely on appearance features or any deep re-identification networks, using only spatial cues and geometric coverage to maintain track consistency.
    
    \item We leverage environment-aware constraints based on road geometry and scene semantics (e.g., roads, crosswalks) to enforce plausible object movement and improve re-identification accuracy.
    
    \item We release a new benchmark dataset for multi-class vehicle and pedestrian tracking, featuring diverse object categories and challenging scenarios such as occlusions and multi-directional traffic flow in CCTV view.

    \item Our method achieves strong performance on the MOT16, MOT17, and MOT20 test sets, with HOTA scores of 66.0, 66.4, and 65.7 respectively—outperforming most state-of-the-art trackers in terms of accuracy.
\end{itemize}

\section{Related Work}
\label{sec:rel_Works}

In the tracking-by-detection paradigm~\cite{wojke2017simple, zhang2022bytetrack, ghalamzan2021deep, hashempour2020data}, object detections—typically generated by deep convolutional networks~\cite{sun2021sparse, cai2018cascade}—are first obtained and then associated across frames to form trajectories. Many frameworks compute similarity between detections and existing tracklets using a combination of geometric and appearance-based cues. While appearance features have been widely adopted, especially through deep learning methods such as DeepSORT~\cite{wojke2017simple} and FeatureSORT \cite{hashempoor2024featuresort} or their extensions employing state space models (SSM) \cite{hashempoorikderi2024gated} such as \cite{wang2024trackingmamba}, they require additional re-identification (Re-ID) networks which significantly increase computational load and memory usage. This added complexity often limits their suitability for real-time or resource-constrained applications.

In contrast, several lightweight tracking methods avoid deep appearance models and instead rely on hand-crafted or geometry-based features, especially in multi-class or traffic-heavy environments where speed and scalability are essential. One of the most notable examples is SORT~\cite{bewley2016simple}, which uses only Kalman filtering and motion-based association via the Hungarian algorithm, offering impressive speed and simplicity. Recent works have shown renewed interest in such efficient designs. For instance, OC-SORT~\cite{cao2023observation} enhances traditional motion-only tracking with improved observation consistency modeling, while BoT-SORT~\cite{aharon2022bot} introduces a modular framework that decouples tracking logic from heavy feature extraction. These approaches demonstrate that robust tracking can be achieved without deep appearance embeddings, particularly when runtime efficiency is a priority.

For occlusion handling, while efficient methods such as ByteTrack address partial occlusions by associating low-confidence detections, their performance degrades in highly crowded scenes where occluded detections become ambiguous and frequent. To improve robustness, many recent approaches integrate appearance-based re-identification features using CNNs \cite{zhang2023motrv2}, which enable better identity recovery after occlusion. However, these methods incur substantial computational overhead, making them less suitable for real-time or resource-constrained applications. More lightweight alternatives like PD-SORT \cite{wang2025pd} and SparseTrack \cite{liu2025sparsetrack} adopt purely geometry-driven strategies to handle occlusion, using pseudo-depth cues derived from 2D bounding boxes. Although these models avoid CNN-based features, they rely on camera viewpoint assumptions and simple depth heuristics that may fail under non-ideal perspectives, perspective distortion, or rapid scene changes. Moreover, depth-based cascaded association in such methods can struggle when targets have similar depths or when long-term occlusion causes identity drift. We instead propose an occlusion-handling mechanism that does not rely on any specific camera viewpoint assumptions and avoids enforcing depth-based cascaded matching, thereby mitigating the limitations associated with perspective distortion and ambiguous spatial proximity in crowded scenes.

\begin{figure*}[t]
\vspace{-0.7cm}
\centering
\includegraphics[height=3.5cm, width=18cm]{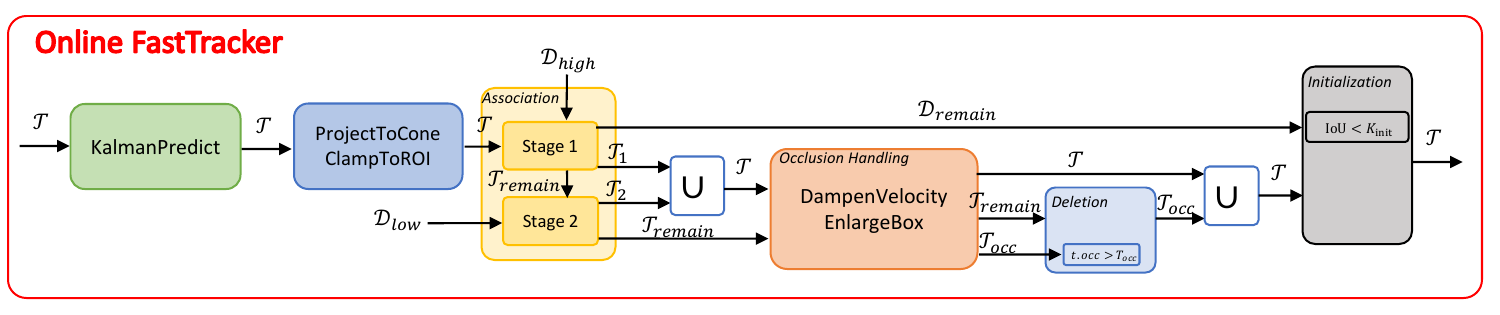}
\caption{ Overall pipeline of Online FastTracker. 
Kalman prediction is refined with region constraints (\texttt{ProjectToCone}, \texttt{ClampToROI}), 
followed by two-stage association using high- and low-confidence detections. 
Occlusion handling (\texttt{DampenVelocity}, \texttt{EnlargeBox}), deletion, and initialization 
ensure stable and consistent tracklet management.}
\setlength{\intextsep}{-10pt}
\vskip -0.2in
\label{fig:fasttracker}
\end{figure*}

Prior works have explored the use of environmental context to support object tracking, though typically in limited or indirect ways. For example, \cite{sani2024graph} and MENet \cite{huang2023menet} employ predefined Regions of Interest (ROIs) to restrict tracking to drivable areas, effectively discarding detections outside these zones. However, these methods do not correct or adapt tracklets that drift beyond the ROIs due to noise or occlusion. In the domain of multi-camera tracking, many approaches leverage contextual zones—such as entry and exit regions—to guide inter-camera association (e.g. Chai et al. \cite{chai2024domain}), but such strategies generally do not make use of fine-grained, single-camera environmental layout to refine trajectories. In contrast, our method incorporates high-level scene structures such as road directions, two-way traffic, and pedestrian crosswalks to adapt tracklets that violate feasible motion patterns. This environment-aware correction is geometry-driven, lightweight, and operates without relying on CNN-based modules, enabling improved tracking consistency and reduced ID switches in urban scenes with complex layout constraints.

Several benchmarks have been introduced for vehicle or multi-class object tracking, but each comes with limitations that leave room for improvement in real-world urban surveillance contexts. The Waymo Open Dataset\cite{sun2020scalability} provides large-scale LiDAR and camera data across various cities for autonomous driving, supporting multiple vehicle classes with high-quality labels. However, it mainly focuses on front-facing vehicle-mounted views and highway scenarios, which differ significantly from urban CCTV viewpoints. KITTI Tracking\cite{geiger2012we} is another widely used benchmark with annotations for cars, pedestrians, and cyclists, but it remains limited in terms of scene diversity of classes and scale, especially under crowded or occluded urban intersections. LMOT\cite{wang2024multi} introduces a challenging nighttime benchmark with low-light conditions, but it lacks multi-class vehicle tracking and other lighting conditions. VETRA\cite{hellekes2024vetra} provides aerial vehicle tracking data that introduces scale variation and perspective distortion, yet it is limited to single-class tracking and overhead viewpoints, which are less relevant to ground-based surveillance systems.
In contrast, our proposed benchmark—although moderate in size—targets multi-class tracking with a strong emphasis on urban CCTV viewpoints, which are underrepresented in existing datasets. It includes diverse object categories, varied lighting conditions, and challenging occlusion scenarios such as dense intersections and bidirectional traffic. This makes it a more realistic and practical resource for developing and evaluating multi-object trackers in city-scale surveillance applications.

\section{Background}
\label{sec:background}
In tracking-by-detection frameworks such as ByteTrack \cite{zhang2022bytetrack}, data association is performed in two stages to balance recall and precision. Let $\mathcal{T}$ denote the set of active tracklets at the current frame, and let the detector provide two groups of detections: $\mathcal{D}_{\mathrm{high}}$ for high-confidence detections (above threshold $\tau_{\mathrm{high}}$) and $\mathcal{D}_{\mathrm{low}}$ for low-confidence detections (between $\tau_{\mathrm{low}}$ and $\tau_{\mathrm{high}}$). In the first stage, $\mathcal{D}_{\mathrm{high}}$ is matched with $\mathcal{T}$ using IoU-based similarity, yielding matched pairs $(\mathcal{T}_1, \mathcal{D}_1)$. Unmatched tracklets and detections are carried forward as $\mathcal{T}_{\mathrm{remain}}$ and $\mathcal{D}_{\mathrm{remain}}$, respectively. In the second stage, the unmatched tracklets $\mathcal{T}_{\mathrm{remain}}$ are further matched with $\mathcal{D}_{\mathrm{low}}$, producing additional associations $(\mathcal{T}_2, \mathcal{D}_2)$. This cascaded design ensures that reliable high-confidence detections are prioritized, while low-confidence detections are selectively used to recover missed targets without introducing excessive false positives.

\section{Approach}

To improve data association in multi-object tracking, we build upon the two-stage matching strategy introduced in ByteTrack (see Background \ref{sec:background}). In this scheme, detections are divided into high-confidence and low-confidence groups: the former provide reliable initial associations, while the latter are selectively used to recover missed targets. This design enhances recall in crowded or ambiguous scenes without substantially increasing false positives. Throughout the paper, we follow the variable notation established in the Background, and introduce additional variables when describing our own extensions.

While effective, this two-stage association alone is insufficient under severe occlusion and dense traffic, where matching ambiguities frequently lead to identity switches or drift. Moreover, the naive initialization and deletion policies used in the baseline and ByteTrack can further degrade performance, highlighting the need for more selective strategies. To address these shortcomings, our framework augments the baseline with three key components: (1) explicit occlusion detection metrics based on spatial overlap, allowing occluded targets to be identified and handled even without detections; (2) environment-aware constraints—such as road directionality, street layout, and pedestrian zones—that refine motion trajectories and prevent implausible movements; and (3) revised initialization and deletion policies that reduce spurious identities while improving recovery of true targets. Together, these additions significantly improve robustness in complex, multi-class urban scenes. The complete FastTrack algorithm is summarized in Figure \ref{fig:fasttracker} and algorithm below.

\begin{algorithm}[t]
\caption{FastTracker}
\label{alg:fasttrack_alg}
\begin{algorithmic}[1]
\Require Video $V$; detector $\mathrm{Det}$; thresholds $\tau_{\mathrm{high}}>\tau_{\mathrm{low}}$;
environment map $\mathcal{M}$;
occlusion params $\mathrm{CP}_{\min}$, $T_{\mathrm{occ}}$;
init/deletion params $K_{\mathrm{init}}$
\Ensure Tracks $\mathcal{T}$
\State $\mathcal{T}\gets \emptyset$

\For{each frame $f_k$ in $V$}
  \State $\mathcal{D}_{\mathrm{high}}, \mathcal{D}_{\mathrm{low}} \gets \mathrm{Det}(f_k)$
  \Statex \quad \Comment{Prediction with map-based constraints}
  \For{each $t\in\mathcal{T}$}
    \State $\hat{x}_t \gets \mathrm{KalmanPredict}(t)$
    \State $R \gets \mathrm{RegionLookup}(\mathcal{M}, \hat{x}_t)$
    \If{$R\neq\varnothing$}
    \Statex \quad \quad \quad \quad \quad\Comment{Direction constraint}
      \State $\hat{x}_t \gets \mathrm{ProjectToCone}(\hat{x}_t, \mathrm{cone}(R))$ 
      \Statex \quad \quad \quad \quad \quad \Comment{Stay inside ROI}
      \State $\hat{x}_t \gets \mathrm{ClampToROI}(\hat{x}_t, R)$ 
    \EndIf
  \EndFor

  \Statex \quad \Comment{Stage 1: Associate with high-confidence detections}
  \State $(\mathcal{T}_1, \mathcal{D}_1) \gets \mathrm{Associate}(\mathcal{T}, \mathcal{D}_{\mathrm{high}})$
  \State $\mathcal{T}_{\mathrm{remain}} \gets \mathcal{T} \setminus \mathcal{T}_1$
  \State $\mathcal{D}_{\mathrm{remain}} \gets \mathcal{D}_{\mathrm{high}} \setminus \mathcal{D}_1$

  \Statex \quad \Comment{Stage 2: Associate with low-confidence detections}
  \State $(\mathcal{T}_2, \mathcal{D}_2) \gets \mathrm{Associate}(\mathcal{T}_{\mathrm{remain}}, \mathcal{D}_{\mathrm{low}})$
  \State $\mathcal{T}_{\mathrm{remain}} \gets \mathcal{T}_{\mathrm{remain}} \setminus \mathcal{T}_2$
  \State $\mathcal{T} \gets \mathcal{T}_1 \cup \mathcal{T}_2$

  \Statex \quad \Comment{Occlusion handling}
  \State $\mathcal{T}_{\mathrm{occ}} \gets \{t \in \mathcal{T}_{\mathrm{remain}} \mid \exists\, t' \in \mathcal{T},\, \mathrm{CP}(t,t') \ge \mathrm{CP}_{\min}\}$
  \For{each $t \in \mathcal{T}_{\mathrm{occ}}$}
    \State $t.\mathrm{occluded} \gets \mathrm{true}$; 
    \State $\mathrm{DampenVelocity}(t)$
    \State $\mathrm{EnlargeBox}(t)$
  \EndFor
  \State $\mathcal{T}_{\mathrm{remain}} \gets \mathcal{T}_{\mathrm{remain}} \setminus \mathcal{T}_{\mathrm{occ}}$

  \Statex \quad \Comment{Deletion policy}
  \State delete tracks in $\mathcal{T}_{\mathrm{remain}}$
  \State delete $t \in \mathcal{T}_{\mathrm{occ}}$ if $t.\mathrm{occ} > T_{\mathrm{occ}}$
  \State $\mathcal{T} \gets \mathcal{T} \cup \mathcal{T}_{\mathrm{occ}}$

  \Statex \quad \Comment{Tracklet initialization}
  \State $\mathcal{D}_{\mathrm{init}} \gets \{d \in \mathcal{D}_{\mathrm{remain}} \mid \max_{t \in \mathcal{T}} \mathrm{IoU}(d, t) < K_{\mathrm{init}}\}$
  \State $\mathcal{T} \gets \mathcal{T} \cup \mathrm{InitializeTracks}(\mathcal{D}_{\mathrm{init}})$
\EndFor
\State \Return $\mathcal{T}$
\end{algorithmic}
\end{algorithm}

\textbf{Motion Prediction.} \hspace{0.2cm}
For each tracklet $t \in \mathcal{T}$, we estimate its future state $\hat{x}_t$ using a class-aware Kalman filter, $\mathrm{KalmanPredict}(t)$. The motion model parameters are selected based on the object class: vehicles such as cars or motorcycles are allowed higher velocities and acceleration bounds, whereas pedestrians are modeled with smoother and slower dynamics. This enables more realistic trajectory predictions, especially in cases of temporary occlusion or detection dropout.

\begin{figure}[h]
    \centering
    \includegraphics[width=\linewidth, keepaspectratio]{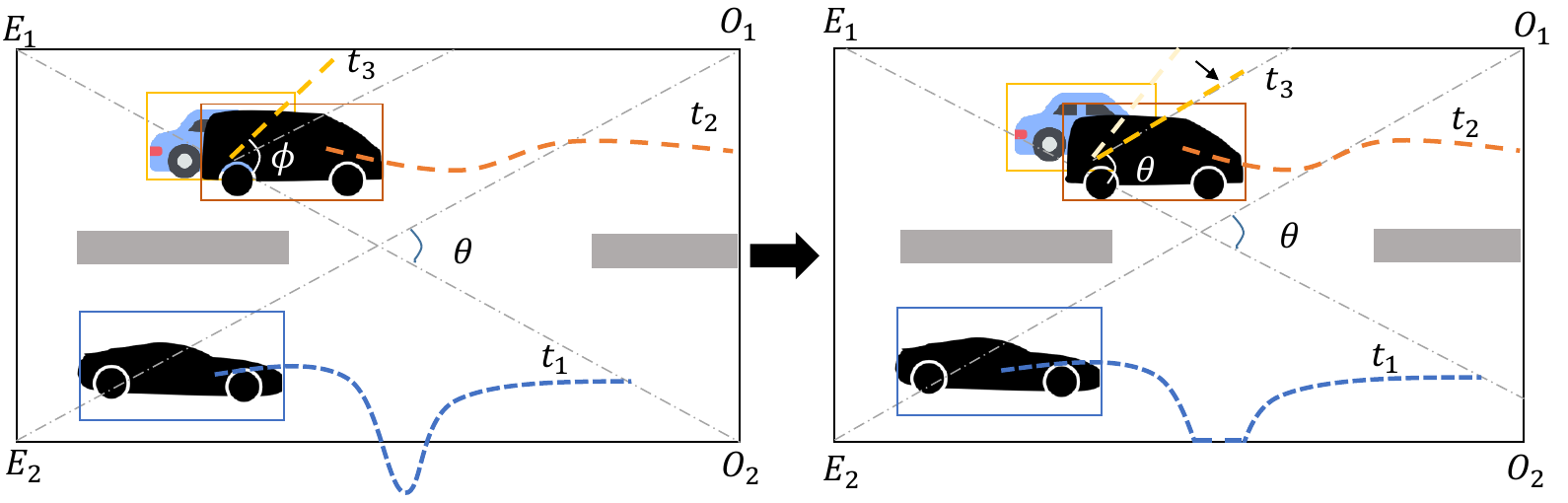}
    \caption{
    Illustration of \texttt{ClampToROI} and \texttt{ProjectToCone}. 
Tracklet $t_1$ drifts outside the region and is clamped back inside. 
Tracklet $t_3$ is misled by occlusion from $t_2$, causing its motion angle $\phi$ to exceed the allowed threshold $\theta$; we project it back within the motion cone.
    }
    \label{fig:roi_cone}
\end{figure}

\textbf{Direction and ROI Constraint.} \hspace{0.2cm}
To prevent tracklets from drifting into implausible directions, we retrieve a region $R$ around the predicted state $\hat{x}_t$ via a region lookup on the environment map $\mathcal{M}$. Each region encodes semantic layout (e.g., drivable roads, pedestrian paths) and is associated with a motion cone that specifies the allowed directionality. 
The cone’s opening angle $\theta$ is derived from the polygonal structure of $R$: given an entrance edge and an exit edge (defined by prior scene-flow knowledge), we construct the two crossing diagonals that connect opposite endpoints of the entrance and exit edges. Let the entrance edge endpoints are $E_1,E_2$ and the exit edge endpoints are $O_1,O_2$, with diagonals $\overrightarrow{E_1O_2}$ and $\overrightarrow{E_2O_1}$, then
\begin{equation}
\theta \;=\; \arccos\!\left(\frac{(\,O_2 - E_1\,)\cdot(\,O_1 - E_2\,)}{\lVert O_2 - E_1\rVert\,\lVert O_1 - E_2\rVert}\right).
\end{equation}

For each active tracklet $t$, we also compute its instantaneous motion direction $\phi$. Let the track center positions at frames $k$ and $k-N$ be $p_k=(x_k,y_k)$ and $p_{k-N}=(x_{k-N},y_{k-N})$. The displacement $\Delta p = p_k - p_{k-N} = (\Delta x, \Delta y)$ gives
\begin{equation}
\phi \;=\; \mathrm{arctan}(2\Delta y,\Delta x).
\end{equation}
The function $\mathrm{ProjectToCone}(\hat{x}_t,\mathrm{cone}(R))$ then projects the predicted position into the direction-constrained cone (i.e., enforces $\phi \in [-\theta/2,\theta/2]$ around the region’s dominant flow), and $\mathrm{ClampToROI}(\hat{x}_t,R)$ keeps the state inside the region. We illustrate these concepts in Figure~\ref{fig:roi_cone} for a one-way road; the same construction extends to two-way roads and crosswalks.

\textbf{Association.}  \hspace{0.2cm}
 In the first stage, high-confidence detections $\mathcal{D}_{\mathrm{high}}$ are matched with active tracklets $\mathcal{T}$ using IoU-based association, yielding matches $(\mathcal{T}_1, \mathcal{D}_1)$. Unmatched tracklets and low score detections, $\mathcal{T}_{\mathrm{remain}}$ and $\mathcal{D}_{\mathrm{low}}$, are then passed to the second stage, where $\mathcal{T}_{\mathrm{remain}}$ is matched with low-confidence detections $\mathcal{D}_{\mathrm{low}}$ to recover hard cases $\mathcal{T}_2$ and we seclude them from $\mathcal{T}_{\mathrm{remain}}$. Finally, the updated active set becomes $\mathcal{T} = \mathcal{T}_1 \cup \mathcal{T}_2$.

\begin{figure}[h]
    \centering
    \includegraphics[width=\linewidth, keepaspectratio]{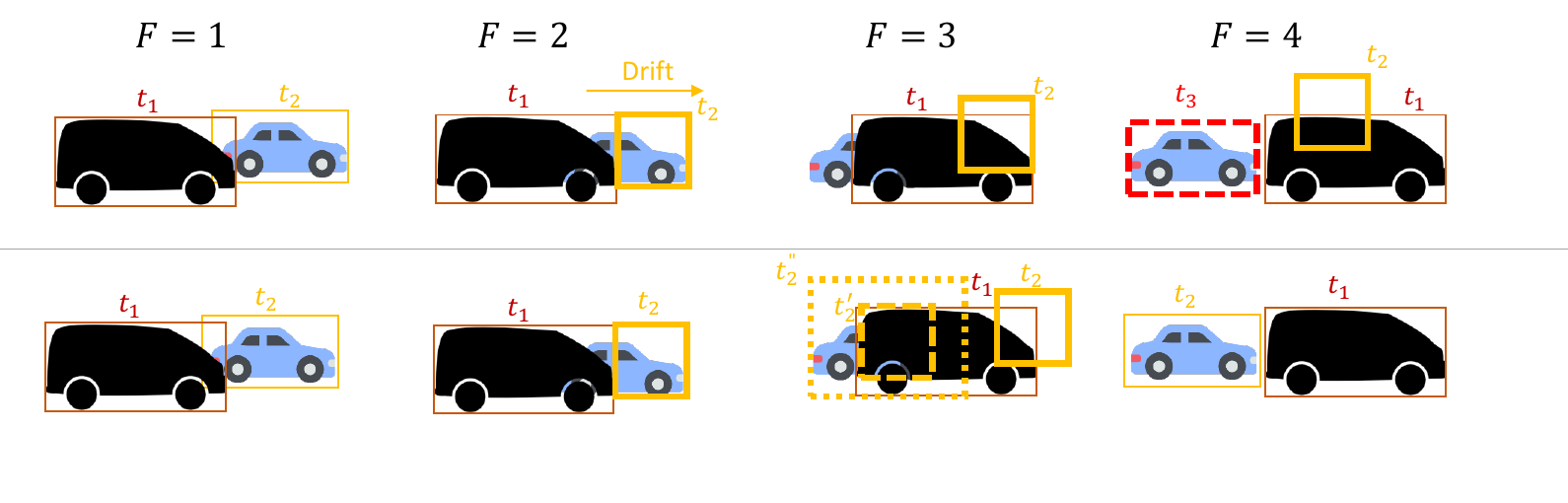}
    \caption{
Illustration of \texttt{DampenVelocity} and \texttt{EnlargeBox} over four frames ($F=1\!-\!4$). 
Top: standard tracking, where fast motion and tight boxes cause drift and ID switches. 
Bottom: our method modifies $t_2$, where $t^{'}_2 \leftarrow t_2$ after \texttt{DampenVelocity} and $t^{''}_2 \leftarrow t^{'}_2$ after \texttt{EnlargeBox}, preserving consistent identities during occlusion.
    }
    \label{fig:occ_alg}
\end{figure}

\textbf{Occlusion Handling.} \hspace{0.2cm}
Unmatched tracklets $\mathcal{T}_{\mathrm{remain}}$ are checked for occlusion by measuring spatial overlap with active tracklets $\mathcal{T}$. If the center-proximity score $\mathrm{CP}(t, t')$, calculated from the IoU, exceeds a threshold $\mathrm{CP}_{\min}$, the tracklet $t$ is considered occluded and added to $\mathcal{T}_{\mathrm{occ}}$. 
For each occluded tracklet, we mark it occluded and apply two corrective operators to stabilize its state during disappearance: (i) velocity damping and (ii) bounding-box enlargement, shown in Figure \ref{fig:occ_alg}. 

For velocity damping, we modify Kalman state of a tracklet represented as 
$x_t $ containing $(p_x,p_y)$ denote the position and $(v_x,v_y)$ the velocity. The operator $\mathrm{DampenVelocity}(t)$ reduces the velocity magnitude by a damping factor $\gamma_{\mathrm{velo}} \in (0,1)$:
\begin{equation}
(v_x, v_y) \;\leftarrow\; \gamma_{\mathrm{velo}} \cdot (v_x, v_y).
\end{equation}
In addition, to mitigate drift during occlusion, the position is softly reset toward its last non-occluded location:
\begin{equation}
    (p_x, p_y) \;\leftarrow\; (p_x, p_y) - \delta_{\mathrm{reset}}.
\end{equation}
This reset constrains unrealistic forward propagation, while subsequent Kalman updates continue from the corrected state.

\begin{figure}[h]
    \centering
    \includegraphics[width=\linewidth, keepaspectratio]{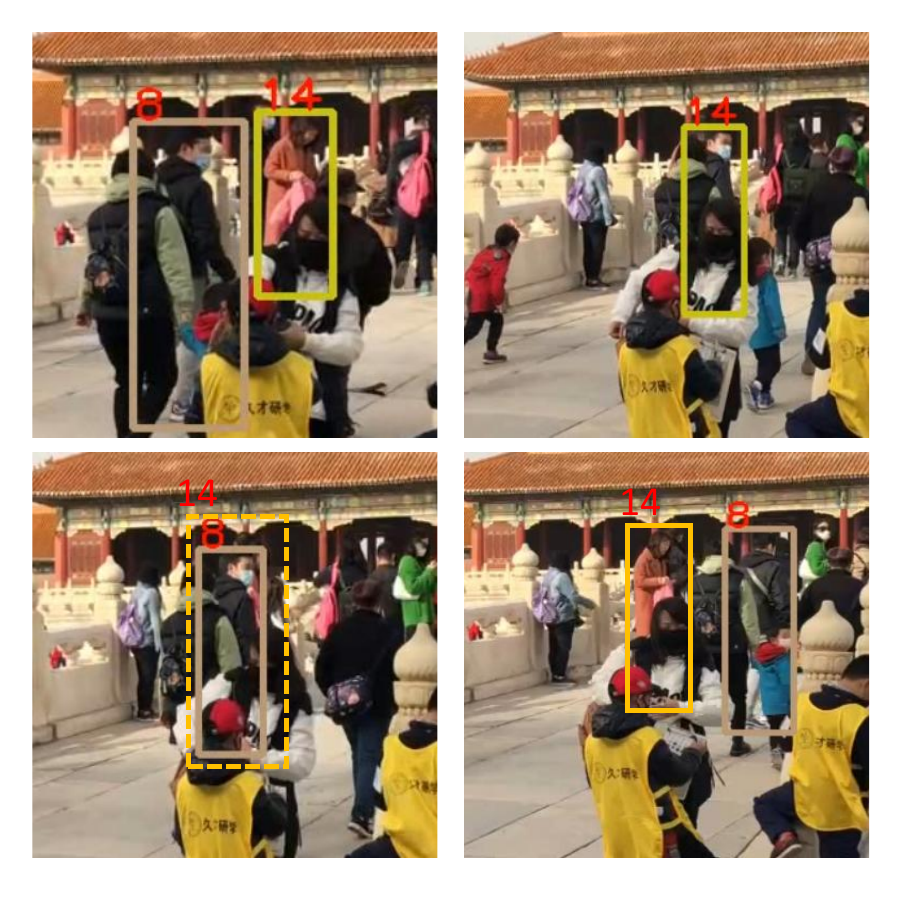}
    \caption{Effectiveness of bounding box enlargement in occlusion handling. Top row: baseline tracker fails to maintain ID during occlusion. Bottom row: enlarged bounding box during occlusion enables successful ID recovery after reappearance.}
    \label{fig:fasttrack_occ_enlarge_bb}
\end{figure}

For bounding-box enlargement, to improve re-identification after reappearance, the bounding box dimensions are slightly enlarged depending on the object class. If $(w,h)$ denote the width and height, then the operator $\mathrm{EnlargeBox}(t)$ applies
\begin{equation}
    (w, h) \;\leftarrow\; \beta_{\mathrm{enlarge}} \cdot (w,h).
\end{equation}
This adjustment increases tolerance for detector noise and partial visibility when the occluded object becomes visible again.
Occluded tracklets $\mathcal{T}_{\mathrm{occ}}$ are then excluded from further association until reappearance, ensuring that identity drift is minimized. The effectiveness of velocity damping and bounding-box enlargement is illustrated in Figure~\ref{fig:fasttrack_occ_enlarge_bb}.

\textbf{Tracklet Initialization and Deletion.} \hspace{0.2cm}
To maintain a clean and reliable set of tracklets, we apply explicit initialization and deletion policies. Remaining high-confidence detections are considered for initialization only if they have low overlap with existing tracklets, i.e., $\max_{t \in \mathcal{T}} \mathrm{IoU}(d, t) < K_{\mathrm{init}}$, ensuring that redundant or duplicate tracks are avoided. On the other hand, unmatched tracklets in $\mathcal{T}_{\mathrm{remain}}$ are deleted unless they are marked as occluded. For occluded tracklets, we allow temporary persistence but delete them if their occlusion age exceeds a threshold $T_{\mathrm{occ}}$. This strategy ensures long-term robustness while avoiding stale or spurious tracks.

\textbf{Post Processing} \hspace{0.2cm}
While our method is designed for fully online inference with minimal reliance on post-processing, we also incorporate two complementary post-processing techniques to showcase their potential benefits. First, global linking is used to associate fragmented tracklets by leveraging spatiotemporal consistency and appearance features extracted via GIAOTracker’s \cite{du2021giaotracker} ResNet50-TP encoder, with tracklet-level matching based on cosine similarity. Second, Gaussian Smoothing Process (GSP) \cite{schulz2018tutorial} is applied to refine tracklet trajectories by modeling nonlinear motion over time. Unlike linear interpolation, GSP incorporates both past and future observations, offering more robust handling of missing detections and smoother trajectory corrections.

\section{Benchmark}

\begin{figure}[t]
    \centering
    \includegraphics[width=\linewidth, keepaspectratio]{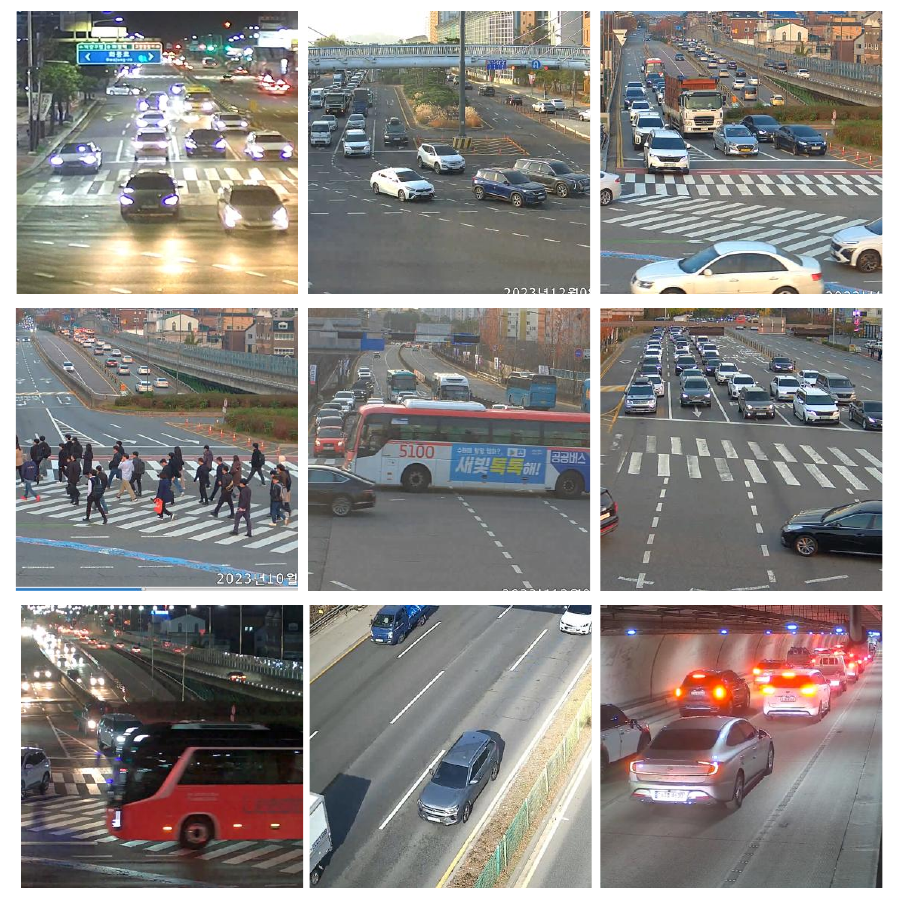}
    \caption{Sample frames from the FastTrack benchmark, showcasing its diverse traffic scenarios including intersections, crosswalks, tunnels, highways, and dense urban roads under varying lighting conditions. The dataset captures complex interactions, high object density, and frequent occlusions.}
    \label{fig:fasttrack_benchmark}
\end{figure}

To comprehensively evaluate multi-object tracking in complex traffic scenes, we introduce the FastTrack benchmark—a diverse and challenging dataset that surpasses existing benchmarks such as UrbanTracker and CityFlow in several key dimensions. FastTrack contains 800K annotated detections across 12 videos, each densely populated with an average of 43.5 objects per frame—more than 5× that of UrbanTracker and over 5× CityFlow—making it particularly suitable for evaluating trackers under extreme crowding and interaction. The dataset spans 9 traffic-related classes, expanding the label diversity beyond those in prior datasets. Furthermore, FastTrack encompasses 12 distinct traffic scenarios, including multilane intersections, crosswalks, tunnels, and merging roads, under varied lighting conditions such as daylight, night scenes, and strong shadow transitions. These factors introduce frequent and severe occlusions, challenging trackers to maintain identity continuity even during long-term disappearances. Compared to existing datasets, which often feature limited scene types and low object density, FastTrack provides a much more realistic and exhaustive benchmark for modern tracking algorithms, especially those designed for deployment in urban traffic environments.
Benchmark statistics and visualizations are provided in Table \ref{tab:dataset_comparison} and Figure \ref{fig:fasttrack_benchmark} respectively.

\begin{table}[t]
\centering
\begin{tabular}{lcccc}
\toprule
\textbf{Attribute} & \textbf{UrbanTracker}  & \textbf{CityFlow} & \textbf{FastTracker} \\
\midrule
Year & 2014 & 2022 & 2025 \\
Detections & 12.5K & 890K & 800k \\
\#Videos & 5 & 40 & 12 \\
Obj./Frame & 5.4 & 8.2 & 43.5 \\
\#Classes & 3 & 1 & 9 \\
\#Scenarios & 1 & 4 & 12 \\
\bottomrule
\end{tabular}
\caption{Comparison of traffic multi-object tracking datasets. ‘Obj./Frame’ denotes the average number of objects per frame.}
\label{tab:dataset_comparison}
\end{table}

\section{Experimental Setup}

\textbf{Datasets. } \hspace{0.3cm} 
For our experiments, we utilized four datasets: MOT16, MOT17, MOT20  \cite{milan2016mot16, dendorfer2020mot20} and our introduced benchmark dataset. MOT16 and MOT17 include a wide range of pedestrian tracking scenarios with both static and moving cameras, where MOT17 further incorporates multiple detector outputs for robust evaluation. MOT20 presents more challenging scenes with extremely crowded environments and heavy occlusions, testing the limits of detection and tracking performance. In addition to these, our custom benchmark introduces even more extreme conditions, featuring massive pedestrian-car crowds, frequent and prolonged occlusions, and visually cluttered scenes. These properties result in significant overlaps between individuals, pushing beyond the visual complexity of the existing MOT datasets and offering a valuable testbed for evaluating the real-world robustness of tracking algorithms.

\textbf{Metrics.}  \hspace{0.3cm}
We evaluate tracking performance using a combination of established metrics. These include the CLEAR metrics \cite{bernardin2008evaluating}—such as MOTA, false positives (FP), false negatives (FN), and identity switches (IDs)—as well as IDF1 \cite{ristani2016performance} and the more recent HOTA metric \cite{luiten2021hota}. While MOTA offers a general measure of tracking accuracy, IDF1 focuses on the quality of identity association. HOTA provides a balanced evaluation by jointly accounting for detection accuracy, association consistency, and localization precision.

\textbf{Implementation Details.} \hspace{0.3cm} 
 For detection, we adopt YOLOX due to its effective balance between speed and accuracy. The detector’s classification and localization heads are trained following the best practices established in prior work \cite{zhang2022bytetrack, du2023strongsort}. At inference time, we apply a non-maximum suppression (NMS) threshold of 0.75. Tracklet association uses an IoU threshold of 0.5, and exponential moving average (EMA) smoothing is applied with a momentum coefficient $\alpha$ of 0.8. We set $\mathrm{CP}_{\min}$ to 0.7, the initialization overlap threshold $K_{init}$ to 0.8, and the occlusion tolerance window $T_{occ}$ to 30 frames. Detection confidence thresholds for classification are $\tau_{low}=0.2$ and $\tau_{high}=0.65$. The region of interest $\mathcal{M}$ is user-configurable, and if direction information is provided, the tracker adapts accordingly. We set $0.75 \leq \gamma_{\mathrm{velo}} \leq 0.9$, $\delta_{\mathrm{reset}} \in \{3,4\}\mathrm{frames}$ and $1.1 \leq \beta_{\mathrm{enlarge}} \leq 1.2$ depending on the object class type.

In post-processing, we limit the maximum gap for Gaussian smoothing interpolation (GSP) to 20 frames. For global linking, we use the MARS ReID dataset \cite{zheng2016mars} for pedestrian training and a vehicle ReID dataset sourced from GIAOTracker for vehicles. It is trained for 60 epochs using Adam optimizer with a cross-entropy loss function and a cosine annealing learning rate schedule. During inference, candidate associations are filtered using a temporal threshold of 15 frames and a spatial distance cap of 70 pixels. Only link scores exceeding 0.9 are accepted. All experiments were conducted on a system equipped with an NVIDIA RTX 4060 GPU with 8GB VRAM. We release our code and benchmark
in public repositories
\footnote{Code and Benchmark: \href{https://github.com/Hamidreza-Hashempoor/FastTracker}{github.com/Hamidreza-Hashempoor/FastTracker}, \href{https://huggingface.co/datasets/Hamidreza-Hashemp/FastTracker-Benchmark}{huggingface.co/datasets/Hamidreza-Hashemp/FastTracker-Benchmark}}
.

\section{Experiment }
To evaluate the effectiveness and robustness of our proposed tracking framework, we conduct extensive experiments across both standard benchmarks and internal studies. First, we perform a comprehensive ablation study to investigate the contribution of each individual component in our pipeline. Then, we present quantitative results on public multi-object tracking benchmarks including MOT16, MOT17, MOT20, and our newly added benchmark dataset, highlighting the competitiveness of our method in both crowded and occluded scenarios.

\subsection{Ablation Studies:}
To better understand the impact of each core component in our framework, we perform a series of ablation studies. Specifically, we examine four major aspects: (1) the effect of our deletion and initialization policies on maintaining track consistency, (2) the contribution of our occlusion-aware mechanism in preserving identities through temporary visual loss, (3) the influence of incorporating user-defined ROI and directional constraints (cone-based filtering), (4) the role of post-processing techniques such as global linking and trajectory smoothing, and (5) check the performance of FastTracker on lighter detectors.

\begin{table}[h]
\centering
\resizebox{\linewidth}{!}{
\begin{tabular}{cc|cc|cc|cc}
\toprule
\multicolumn{2}{c|}{} 
& \multicolumn{2}{c|}{\textbf{MOT17-val}} 
& \multicolumn{2}{c|}{\textbf{MOT20-val}} 
& \multicolumn{2}{c}{\textbf{FastTrack}} \\
\textbf{Del} & \textbf{Init} 
& \textbf{MOTA} $\uparrow$ & \textbf{HOTA} $\uparrow$ 
& \textbf{MOTA} $\uparrow$ & \textbf{HOTA} $\uparrow$ 
& \textbf{MOTA} $\uparrow$ & \textbf{HOTA} $\uparrow$ \\
\midrule
\checkmark &            &    79.4   &    63.5   &   74.5    &    62.8   &   60.1    &   57.2    \\
           & \checkmark &    79.6   &    63.7   &    74.6   &    63.0   &   60.4    & 57.6      \\
           \rowcolor{gray!20} 
\checkmark & \checkmark &    \textbf{79.9}   &    \textbf{64.0}   &    \textbf{75.2}   &     \textbf{63.3}  &    \textbf{60.9}   &   \textbf{58.0}    \\
\bottomrule
\end{tabular}
}
\caption{Ablation study on deletion and initialization policies across three datasets.}
\label{tab:del_ini_ablation}
\end{table}

\textbf{Deletion and Initialization Policies:}
As shown in Table~\ref{tab:del_ini_ablation}, enabling either deletion or initialization independently brings modest gains, while their combination consistently yields the best performance. Specifically, on MOT17, jointly applying both strategies improves MOTA from 79.4 to 79.9 and HOTA from 63.5 to 64.0, resulting in absolute gains of +0.5 MOTA and +0.5 HOTA. Similarly, on MOT20, we observe a +0.7 MOTA and +0.5 HOTA improvement. For the more challenging FastTrack dataset, the joint configuration boosts MOTA from 60.1 to 60.9 and HOTA from 57.2 to 58.0. These results confirm the effectiveness of our proposed deletion and initialization policies in enhancing both detection reliability and identity preservation.

\begin{table}[h]
\centering
\resizebox{\linewidth}{!}{
\begin{tabular}{c|cc|cc|cc}
\toprule
\multicolumn{1}{c|}{} 
& \multicolumn{2}{c|}{\textbf{MOT17-val}} 
& \multicolumn{2}{c|}{\textbf{MOT20-val}} 
& \multicolumn{2}{c}{\textbf{FastTrack}} \\
\textbf{Occ} 
& \textbf{MOTA} $\uparrow$ & \textbf{HOTA} $\uparrow$ 
& \textbf{MOTA} $\uparrow$ & \textbf{HOTA} $\uparrow$ 
& \textbf{MOTA} $\uparrow$ & \textbf{HOTA} $\uparrow$ \\
\midrule
            & 79.0 & 63.1 & 74.1 & 62.1 & 59.2 & 56.8 \\
            \rowcolor{gray!20} 
\checkmark  & \textbf{80.4} & \textbf{65.2} & \textbf{76.7} & \textbf{64.4} & \textbf{63.4} & \textbf{60.7} \\
\bottomrule
\end{tabular}
}
\caption{Ablation study on occlusion handling.}
\label{tab:occ_ablation}
\end{table}

\textbf{Occlusion Handling:} \hspace{0.2cm}
To isolate the effect of occlusion handling, we disable our proposed deletion and initialization policies and revert to the conventional strategy—removing tracklets immediately when not visible and initializing new ones from all high-confidence detections. As shown in Table~\ref{tab:occ_ablation}, incorporating our occlusion-aware mechanism yields a notable performance gain across all datasets. Specifically, on MOT17, HOTA improves by 2.1 points (from 63.1 to 65.2) and MOTA by 1.4 points (from 79.0 to 80.4). On the more crowded MOT20 dataset, our approach increases HOTA by 2.3 points and MOTA by 2.6 points. The largest gains are observed on FastTrack, where HOTA improves by 3.9 points and MOTA by 4.2 points. These results highlight the critical role of explicit occlusion modeling in maintaining accurate identity association under challenging scenarios.

\textbf{ROI and Direction.} \hspace{0.3cm}
To further refine the association process in structured scenes, we examine the effect of incorporating Region-of-Interest (ROI) filtering and direction (Dir) constraints on the FastTrack dataset, as shown in Table \ref{tab:roi_ablation}. These evaluations are conducted on top of other modules, including the proposed occlusion handling and deletion/initialization policies. Applying the ROI constraint alone achieves a performance of 63.5 MOTA / 60.8 HOTA, while using only the direction constraint slightly improves the scores to 63.6 MOTA / 61.0 HOTA. When both constraints are combined, the method reaches the best result of 63.8 MOTA / 61.0 HOTA, demonstrating modest yet consistent gains. 

\begin{table}[h]
\centering
\resizebox{0.6\linewidth}{!}{ % reduced size to better fit one-column layout
\begin{tabular}{cc|cc}
    \toprule
    \multicolumn{2}{c|}{} 
    & \multicolumn{2}{c}{\textbf{FastTrack}} \\
    \textbf{ROI} & \textbf{Dir} 
    & \textbf{MOTA} $\uparrow$ & \textbf{HOTA} $\uparrow$ \\
    \midrule
    \checkmark &            & 63.7 & 61 \\
               & \checkmark & 64.2 & 61.3 \\
               \rowcolor{gray!20} 
    \checkmark & \checkmark & \textbf{64.4} & \textbf{61.5} \\
    \bottomrule
    \end{tabular}
}
\caption{Ablation study on ROI and direction constraint (Dir) using the FastTrack dataset.}
\label{tab:roi_ablation}
\end{table}

\textbf{Post Processing:}
To evaluate the contribution of the post-processing stage, we analyze the impact of incorporating the GSI,  which outperforms conventional interpolation techniques \cite{hashempoor2025deep}, and global linking (G-Link) on the final tracking performance across three datasets. Importantly, all results reported here are based on the full online system, which includes our proposed deletion/initialization policies and occlusion handling mechanisms. The post-processing modules are applied offline after generating the initial tracks. As shown in Table~\ref{tab:gsp_glink_ablation}, enabling either GSP or G-Link individually brings consistent improvements over the base output, with G-Link slightly outperforming GSP in terms of HOTA across all datasets.

\begin{table}[h]
\centering
\resizebox{\linewidth}{!}{
\begin{tabular}{cc|cc|cc|cc}
    \toprule
    \multicolumn{2}{c|}{} 
    & \multicolumn{2}{c|}{\textbf{MOT17-val}} 
    & \multicolumn{2}{c|}{\textbf{MOT20-val}} 
    & \multicolumn{2}{c}{\textbf{FastTrack}} \\
    \textbf{GSP} & \textbf{G-Link} 
    & \textbf{MOTA} $\uparrow$ & \textbf{HOTA} $\uparrow$ 
    & \textbf{MOTA} $\uparrow$ & \textbf{HOTA} $\uparrow$ 
    & \textbf{MOTA} $\uparrow$ & \textbf{HOTA} $\uparrow$ \\
    \midrule
    \checkmark &            &    82.0   &    66.5   &   80.1    &    65.9   &   64.7    &   61.9    \\
               & \checkmark &    82.0   &    66.6   &    80.1   &    66   &   64.8    & 62      \\
               \rowcolor{gray!20} 
               \checkmark & \checkmark &    \textbf{82.2}   &    \textbf{66.6}   &    \textbf{80.3}   &     \textbf{66.3}  &    \textbf{65}   &   \textbf{62.2
               }    \\
    \bottomrule
    \end{tabular}
}
\caption{Ablation study on global spatial prior (GSP) and global linking (G-Link) across three datasets.}
\label{tab:gsp_glink_ablation}
\end{table}

\textbf{Tracking performance of light modified YOLOX.} \hspace{0.3cm}
FastTracker demonstrates robust performance even when paired with lightweight YOLOX detectors as shown in Table \ref{tab:light_models}. As the model size decreases from YOLOX-L (61M) to YOLOX-Nano (1M), the tracking accuracy degrades gracefully, maintaining strong metrics across the board. Notably, YOLOX-M and YOLOX-S achieve MOTA scores of 78.1 and 74.6 respectively, outperforming several baselines that rely on larger detectors and heavier re-identification modules. Even with YOLOX-Nano, FastTracker achieves a convincible MOTA of 68.3 and IDF1 of 71.2, enabling real-time deployment on edge devices with limited resources. These results highlight the efficiency and scalability of FastTracker across a range of detector capacities.

\begin{table}[h]
    \caption{Performance of FastTracker using lightweight YOLOX models on the MOT17 validation set.}
	\begin{minipage}{1\linewidth} 
		\label{tab:light_models}
		\centering
        \begin{tabular}{c c c c c c}
            \toprule
            \textbf{Detector} & \textbf{Params} & \textbf{MOTA} $\uparrow$ & \textbf{HOTA} $\uparrow$ & \textbf{IDF1} $\uparrow$ & \textbf{IDs} $\downarrow$ \\
            \midrule
            YOLOX-L       & 61M   & 79.3 & 65.1 & 79.8 & 936 \\
            YOLOX-M       & 28M   & 78.1 & 64.0 & 77.6 & 1141 \\
            YOLOX-S       & 10.5M & 74.6 & 61.7 & 74.7 & 1477 \\
            YOLOX-Tiny    & 5.8M  & 73.9 & 60.2 & 74.1 & 1696 \\
            YOLOX-Nano    & 1M    & 68.3 & 56.8 & 71.2 & 2285 \\
            \bottomrule
        \end{tabular}
	\end{minipage}
\end{table}

\subsection{Benchmark Evaluation}
In the final part of our experiments, we evaluate the effectiveness of our proposed tracker against state-of-the-art methods on MOT16, MOT17, MOT20, and FastTrack. To demonstrate our online tracking performance, we submit our tracking outputs (without any post-processing such as GSP or G-Link) to the respective evaluation servers. This setup ensures that all reported results reflect the performance of our online system only, highlighting the strength of our tracking framework under standard benchmark protocols and constraints.

\textbf{MOT16 and MOT17:} \hspace{0.2cm}
Table \ref{tab:mot16} presents the official MOT16 benchmark results. Our method, FastTracker, achieves the highest scores among all compared methods, with MOTA of 79.1 and HOTA of 66.0, surpassing recent state-of-the-art approaches. Compared to FeatureSORT (MOTA 77.9, HOTA 62.8), we improve by +1.2 MOTA and +3.2 HOTA. Similarly, against StrongSORT (MOTA 77.8, HOTA 63.8), we see gains of +1.3 MOTA and +2.2 HOTA. Notably, we also achieve the lowest number of ID switches (290), demonstrating superior identity preservation throughout the sequence.

Table \ref{tab:mot17} shows the performance of FastTracker on the MOT17 benchmark. Our method achieves a new state-of-the-art with MOTA of 81.8 and HOTA of 66.4, clearly outperforming all previous trackers. Compared to FeatureSORT (MOTA 79.6, HOTA 63.0), we achieve improvements of +2.2 MOTA and +3.4 HOTA. Against StrongSORT (MOTA 78.3, HOTA 63.5), the gains are +3.5 MOTA and +2.9 HOTA. Even the widely adopted ByteTrack shows lower performance (MOTA 78.9, HOTA 62.8), with a margin of +2.9 MOTA and +3.6 HOTA. Additionally, FastTracker achieves the lowest number of ID switches (885) and the lowest FN (75162), confirming both strong identity preservation and detection quality.

\begin{table}[h]
    \caption{
            MOT16
        }
	\begin{minipage}{1\linewidth} 
		\label{tab:mot16}
		\centering
		\scriptsize
		        \begin{tabular}{| c | c  ccccc }
                    \toprule
                    \multicolumn{1}{c}{Method} & \multicolumn{1}{c}{MOTA$\uparrow$} & \multicolumn{1}{c}{HOTA$\uparrow$} & \multicolumn{1}{c}{IDF1$\uparrow$} & \multicolumn{1}{c}{FP$\downarrow$} & \multicolumn{1}{c}{FN$\downarrow$} & \multicolumn{1}{c}{IDs$\downarrow$} \\
                    \hline
                    \multicolumn{1}{c}{QDTrack \cite{pang2021quasi}} & 69.8  & 54.5 & 67.1 	 & 9861& 44050& 1097\\
                    \multicolumn{1}{c}{TraDes \cite{wu2021track}} & 70.1 & 53.2 & 64.7 & \textbf{8091} & 45210& 1144 \\
                    \multicolumn{1}{c}{CSTrack \cite{liang2022rethinking}} & 75.6  & 59.8 & 73.3  & 9646& 33777& 1121\\
                    \multicolumn{1}{c}{GSDT \cite{wang2021joint}} & 74.5 & 56.6 & 68.1 & 8913& 36428& 1229 \\
                    \multicolumn{1}{c}{RelationTrack \cite{yu2022relationtrack}} & 75.6 & 61.7 & 75.8 & 9786& 34214& 448 \\
                    \multicolumn{1}{c}{OMC \cite{liang2022one}} & 76.4 & 62
                    9& 74.1 & 10821 & 31044& 1087\\
                    \multicolumn{1}{c}{CorrTracker \cite{wang2021multiple}} & 76.6  & 61.0 & 74.3 & 10860& 30756& 979 \\
                    \multicolumn{1}{c}{SGT \cite{hyun2205detection}} & 76.8  & 61.2 & 73.5 & 10695 & 30394& 1276\\
                    \multicolumn{1}{c}{FairMOT \cite{zhang2021fairmot}} & 74.9 & 58.3 & 72.8 & 9952& 38451& 1074\\
                    \multicolumn{1}{c}{StrongSORT \cite{du2023strongsort}} & 77.8 & 63.8 & 78.3 & 11254 & 32584 & 1538\\
                    \multicolumn{1}{c}{FeatureSORT \cite{hashempoor2024featuresort}} & 77.9  & 62.8 & 76.3 & 14827 & \textbf{26877} & 597\\
                    \hline
                    \hline
                    \rowcolor{gray!20} 
                    \multicolumn{1}{c}{FastTracker} & \textbf{79.1}  & \textbf{66.0} & \textbf{81.0} & 8785 & 29028 & \textbf{290}\\
                    \hline
                \end{tabular}
	\end{minipage}
\end{table}

\begin{table}[h]
\centering
\resizebox{\linewidth}{!}{
\begin{tabular}{| c | c  ccccc }
                    \toprule
                    \multicolumn{1}{c}{Method} & \multicolumn{1}{c}{MOTA$\uparrow$} & \multicolumn{1}{c}{HOTA$\uparrow$} & \multicolumn{1}{c}{IDF1$\uparrow$} & \multicolumn{1}{c}{FP$\downarrow$} & \multicolumn{1}{c}{FN$\downarrow$} & \multicolumn{1}{c}{IDs$\downarrow$} \\
                    \hline
                    \multicolumn{1}{c}{CTracker \cite{peng2020chained}} & 66.6 & 49.0  & 57.4  & 22284 & 160491 & 5529\\
                    \multicolumn{1}{c}{CenterTrack \cite{zhou2020tracking}} & 67.8 & 60.3 & 64.7 & 18489 & 160332 & 3039 \\
                    \multicolumn{1}{c}{QDTrack \cite{pang2021quasi}} & 68.7 & 63.5 & 66.3 & 26598 & 146643 & 3378\\
                    \multicolumn{1}{c}{TraDes \cite{wu2021track} } & 69.1 & 52.7  & 63.9 & 20892 & 150060 & 3555 \\
                    \multicolumn{1}{c}{SOTMOT \cite{zheng2021improving} } & 71 & 64.1 & 71.9 & 39537 & 118983 & 5184 \\
                    \multicolumn{1}{c}{GSDT \cite{wang2021joint} } & 73.2 & 55.2 & 66.5 & 26397 & 120666 & 3891\\
                    \multicolumn{1}{c}{RelationTrack \cite{yu2022relationtrack}} & 73.8 & 59.9 & 74.7 & 27999 & 118623 & 1374 \\
                    \multicolumn{1}{c}{TransTrack \cite{sun2020transtrack}} & 74.5 & 54.1 & 63.9 & 28323 & 112137 & 3663 \\
                    \multicolumn{1}{c}{OMC–F \cite{liang2022one}} & 74.7 & 56.8 & 73.8 & 30162 & 108556 & -\\
                    \multicolumn{1}{c}{CSTrack \cite{liang2022rethinking}} & 74.9 & 59.3 & 72.3 & 23847 & 114303 & 3567\\
                    \multicolumn{1}{c}{OMC \cite{liang2022one}} & 76 & 57.1 & 73.8 & 28894 & 101022 & -\\
                    \multicolumn{1}{c}{SGT \cite{hyun2205detection}} & 76.3 & 57.3 & 72.8 & 25974 & 102885 & 4101\\
                    \multicolumn{1}{c}{CorrTracker \cite{wang2021multiple}} & 76.4 & 58.4 & 73.6 & 29808 & 99510 & 3369 \\
                    \multicolumn{1}{c}{FairMOT \cite{zhang2021fairmot} } & 73.7 & 59.3  & 72.3 & 27507 & 117477 & 3303\\
                    \multicolumn{1}{c}{DeepSORT \cite{wojke2017simple} } & 78 & 61.2  & 74.5 & 29852 & 94716 & 1821\\
                    \multicolumn{1}{c}{ByteTrack \cite{zhang2022bytetrack}} & 78.9 & 62.8 & 77.2 & 25491 & 83721 & 2196 \\
                    \multicolumn{1}{c}{StrongSORT \cite{du2023strongsort}} & 78.3 & 63.5 & 78.5 & 27876 & 86205 & 1446 \\
                    \multicolumn{1}{c}{FeatureSORT \cite{hashempoor2024featuresort}} & 79.6  & 63 & 77.2 & 29588 & 83132 & 2269\\
                    \multicolumn{1}{c}{ FLWM~\cite{liu2024iof}} & 80.5 & 64.9 & 79.9 & 27245 & 81653 & 1370 \\
                    \multicolumn{1}{c}{ SparseTrack~\cite{liu2025sparsetrack}} & \textcolor{red}{81} & 65.1 & 80.1 & 23904 & 81927 & 1170 \\
                    \multicolumn{1}{c}{ PD-SORT~\cite{wang2025pd}} & 79.3 & 63.9 & 79.2 & \textcolor{red}{17028} & 101130 & 1062 \\
                    \multicolumn{1}{c}{ C-TWiX~\cite{miah2025learning}} & 78.1 & 63.1 & 76.3 & 20964 & 96642 & 5820 \\
                    \multicolumn{1}{c}{ TOPICTrack~\cite{cao2025topic}} & 78.8 & 63.9 & 78.6 & \textcolor{blue}{\textbf{17010}} & 101130 & 1515 \\
                    \multicolumn{1}{c}{ EscapeTrack~\cite{yi2025escapetrack}} & 80.8 & \textcolor{red}{66.2} & \textcolor{red}{81.9} & 34908 & \textcolor{red}{75252} & \textcolor{red}{1061} \\
                    \hline
                    \hline
                    \rowcolor{gray!20} 
                    \multicolumn{1}{c}{FastTracker} & \textcolor{blue}{\textbf{81.8}}  & \textcolor{blue}{\textbf{66.4}} & \textcolor{blue}{\textbf{82.0}} & 26850 & \textcolor{blue}{\textbf{75162}} & \textcolor{blue}{\textbf{885}}\\
                    \hline
                \end{tabular}
}
\caption{MOT17. Best: \textcolor{blue}{\textbf{Bold }blue}, Second best: \textcolor{red}{Red}.}
\label{tab:mot17}
\end{table}

\textbf{MOT20:} \hspace{0.3cm}
Table \ref{tab:mot20} presents the results on the challenging MOT20 benchmark. FastTracker achieves the highest performance across most key metrics, setting a new state-of-the-art with MOTA 77.9, HOTA 65.7, and IDF1 81.0. This represents a significant improvement over prior methods. Compared to FeatureSORT (MOTA 76.6, HOTA 61.3, IDF1 75.1), we see gains of +1.3 MOTA, +4.4 HOTA, and +5.9 IDF1. Against ByteTrack (MOTA 75.7, HOTA 60.9, IDF1 74.9), our method improves by +2.2 MOTA, +4.8 HOTA, and +6.1 IDF1. Notably, FastTracker also achieves the lowest ID switches (684) among all trackers, indicating robust identity preservation even in extremely crowded scenes. These results confirm that FastTracker delivers state-of-the-art tracking performance in high-density environments relying on its occlusion handling capabilities.

\begin{table}[h]
\centering
\resizebox{1\linewidth}{!}{ % reduced size to better fit one-column layout
\begin{tabular}{| c | c  ccccc }
                    \toprule
                    \multicolumn{1}{c}{Method} & \multicolumn{1}{c}{MOTA$\uparrow$} & \multicolumn{1}{c}{HOTA$\uparrow$} & \multicolumn{1}{c}{IDF1$\uparrow$} & \multicolumn{1}{c}{FP$\downarrow$} & \multicolumn{1}{c}{FN$\downarrow$} & \multicolumn{1}{c}{IDs$\downarrow$} \\
                    \hline
                    \multicolumn{1}{c}{SORT \cite{bewley2016simple}} & 42.7 & 36.1 & 45.1 & 28398 & 287582 & 4852 \\
                    \multicolumn{1}{c}{Tracktor++ \cite{bergmann2019tracking} } & 52.6 & 42.1 & 52.7 & 35536 & 236680 &  1648\\
                    \multicolumn{1}{c}{CSTrack \cite{liang2022rethinking}  } & 66.6 & 54 & 68.6 & 25404 & 144358 & 3196 \\
                    \multicolumn{1}{c}{CrowdTrack \cite{stadler2021performance} } & 70.7 & 55 & 68.2 & 21928 & 126533 & 3198 \\
                    \multicolumn{1}{c}{RelationTrack \cite{yu2022relationtrack}} & 67.2 & 56.5 & 70.5 & 61134 & 104597 & 4243\\
                    \multicolumn{1}{c}{DeepSORT \cite{wojke2017simple}} & 71.8 & 57.1 & 69.6 & 37858 & 101581 & 3754 \\
                    \multicolumn{1}{c}{TransTrack \cite{sun2020transtrack} } & 64.5 & 48.9 & 59.2 & 28566 & 151377 & 3565\\
                    \multicolumn{1}{c}{CorrTracker \cite{wang2021multiple}} & 65.2 & 57.1 & 69.1 & 79429 & 95855 & 5193\\
                    \multicolumn{1}{c}{GSDT \cite{wang2021joint} } & 67.1 & 53.6 & 67.5 & 31913 & 135409 & 3131 \\
                    \multicolumn{1}{c}{SOTMOT \cite{zheng2021improving}} & 68.6 & 55.7 & 71.4 & 57064 & 101154 & 4209
                    \\
                    \multicolumn{1}{c}{OMC \cite{liang2022one} } & 73.1 & 60.5 & 74.4 & \textcolor{red}{16159} & 108654 & 779 
                    \\
                    \multicolumn{1}{c}{FairMOT \cite{zhang2021fairmot} } & 61.8 & 54.6 & 67.3 & 103404 & 88901 & 5243
                    \\
                    \multicolumn{1}{c}{SGT \cite{hyun2205detection}} & 64.5 & 56.9 & 62.7 & 67352 & 111201 & 4909
                    \\
                    \multicolumn{1}{c}{ByteTrack \cite{zhang2022bytetrack} } & 75.7 & 60.9 & 74.9 & 26249 & 87594 & 1223
                    \\
                    \multicolumn{1}{c}{StrongSORT \cite{du2023strongsort}} & 72.2 & 61.5 & 75.9 & 16632 & 117920 & \textcolor{red}{770}
                    \\
                    \multicolumn{1}{c}{FeatureSORT \cite{hashempoor2024featuresort}} & 76.6  & 61.3 & 75.1 & 25083 & 95027 & 1081\\
                    \multicolumn{1}{c}{  FLWM~\cite{liu2024iof}} & 77.7 & 62 & 75 & 25019 & 88959 & 1530 \\
                    \multicolumn{1}{c}{ SparseTrack~\cite{liu2025sparsetrack}} & \textcolor{blue}{\textbf{78.2}} & 63.4 & 77.3 & 25108 & \textcolor{red}{86720} & 1116 \\
                    \multicolumn{1}{c}{ PD-SORT~\cite{wang2025pd}} & 75.4 & 62.6 & 76.7 & 23974 & 93662 & 908 \\
                    \multicolumn{1}{c}{ C-TWiX~\cite{miah2025learning}} & 75 & 62.4 & 74.1 & 25933 & 115240 & 2048 \\
                    \multicolumn{1}{c}{ TOPICTrack~\cite{cao2025topic}} & 72.4 & 62.6 & 77.6 & \textcolor{blue}{\textbf{10986}} & 131088 & 869 \\
                    \multicolumn{1}{c}{ EscapeTrack~\cite{yi2025escapetrack}} & 77.4 & \textcolor{red}{64.8} & \textcolor{red}{80.5} & 30666 & \textcolor{blue}{\textbf{85188}} & 1061 \\
                    \hline
                    \hline
                    \rowcolor{gray!20} 
                    \multicolumn{1}{c}{FastTracker} & \textcolor{red}{77.9}  & \textcolor{blue}{\textbf{65.7}} & \textcolor{blue}{\textbf{81.0}} & 24590 & 89243 & \textcolor{blue}{\textbf{684}} \\
                    \hline
                \end{tabular}
}
\caption{MOT20.  Best: \textcolor{blue}{\textbf{Bold }blue}, Second best: \textcolor{red}{Red}.}
\label{tab:mot20}
\end{table}

\textbf{DanceTrack.} \hspace{0.2 cm}
  We include comparison with the DanceTrack to demonstrate that FastTracker generalizes well beyond the MOT benchmarks. As shown in Table~\ref{tab:dancetrack}, FastTracker achieves the highest scores with 93.4 MOTA and 65.9 HOTA, outperforming prior methods.

\begin{table}[h]
    % \vspace{-0.2cm}
    \centering
    % \tiny
    \vspace{0.1cm}
    \begin{tabular}{l c c c}
        \toprule
        Method & MOTA$\uparrow$ & HOTA$\uparrow$ & IDF1$\uparrow$ \\
        \hline
        TraDes \cite{wu2021track} & 86.2 & 43.3 & 41.2 \\
        SGT \cite{hyun2205detection} & 85.1 & 43.7 & 47.2 \\
        DeepSORT \cite{wojke2017simple} & 88.2 & 46.4 & 51.7 \\
        FairMOT \cite{zhang2021fairmot} & 82.2 & 39.7 & 40.8 \\
        ByteTrack \cite{zhang2022bytetrack} & 89.6 & 47.7 & 53.9 \\
        StrongSORT \cite{du2023strongsort} & 91.1 & 55.6 & 55.2 \\
        FLWM~\cite{liu2024iof} & 90 & 62.2 & 61.4 \\
        UCMCTrack~\cite{yi2024ucmctrack} & 88.9 & \textcolor{red}{65} & 63.6 \\
        SparseTrack~\cite{liu2025sparsetrack} & 91.3 & 55.5 & 58.3 \\
        PD-SORT~\cite{wang2025pd} & 87.4 & 55.5 & 55.4 \\
        C-TWiX~\cite{miah2025learning} & \textcolor{red}{91.4} & 62.2 & 63.5 \\
        TOPICTrack~\cite{cao2025topic} & 89.3 &  55.9 & 54.5 \\
        EscapeTrack~\cite{yi2025escapetrack} & 89.5 & 62 &  \textcolor{red}{66.4} \\
        \hline
        \rowcolor{gray!20}
        FastTracker & \textcolor{blue}{\textbf{93.4}} & \textcolor{blue}{\textbf{65.9}} & \textcolor{blue}{\textbf{67.2}} \\
        \hline
    \end{tabular}
    \caption{Dance: Performance comparison with SOTA. Best: \textcolor{blue}{\textbf{Bold }blue}, Second best: \textcolor{red}{Red}.}
\label{tab:dancetrack}
    % \vskip -0.1in
\end{table}

\begin{figure}[h]
    \centering
    \includegraphics[width=0.9\linewidth, keepaspectratio]{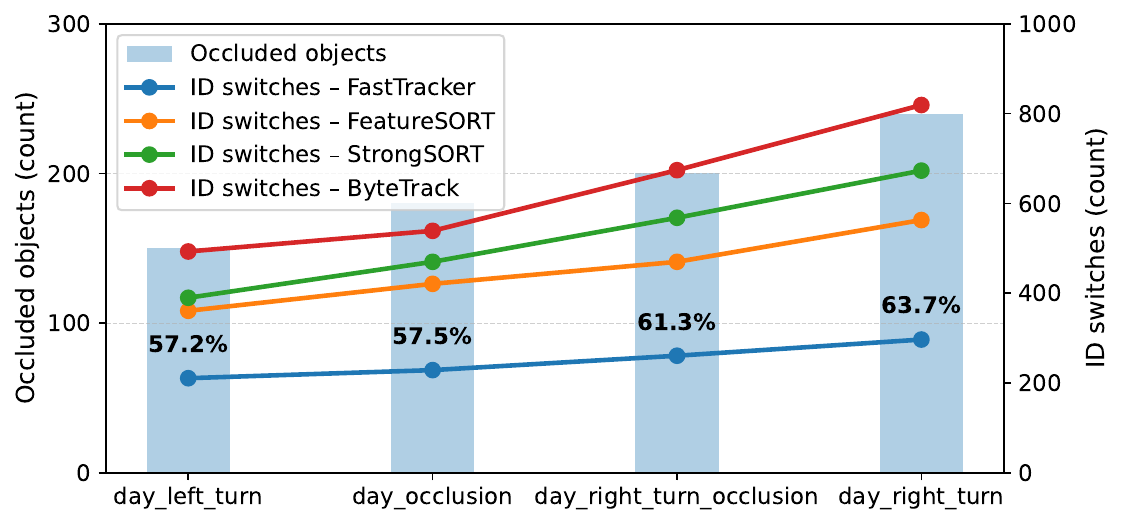}
    \caption{Reduction in ID switches vs.\ baseline. Bars: occluded objects (left axis). Lines: ID switches per method (right axis).
   }
    \label{fig:occlusion_idswitch}
\end{figure}

\textbf{FastTracker Benchmark:} \hspace{0.3cm}
Table \ref{tab:fast_bench} presents the results on the FastTracker Benchmark, a challenging internal benchmark designed to assess robustness in dense tracking scenarios. FastTracker achieves the best MOTA (63.8), IDF1 (79.2) and HOTA (61.0), surpassing FeatureSORT and ByteTrack in association accuracy. It also achieves the lowest number of identity switches (251), demonstrating strong consistency across frames. While FastTracker leads in MOTA (63.8), it also shows a better balance between detection and identity preservation, outperforming StrongSORT and FeatureSORT by +1.3 and +3 HOTA, respectively, and significantly improving IDF1 by up to +1.8. These results highlight FastTracker’s strength in maintaining accurate identity tracking under crowded conditions.
Figure~\ref{fig:occlusion_idswitch} illustrates the relationship between occlusion levels in 4 FastTracker benchmark videos and identity switches. 
The percentages shown in the bars denote the relative reduction in ID switches of FastTracker w.r.t. the baseline ByteTrack, 
indicating the outperformance of FastTracker. 
Notably, higher reductions are observed in videos with more occluded objects, highlighting the robustness of our method under challenging conditions.

\begin{table}[h]
    \caption{
            FastTracker Benchmark
        }
	\begin{minipage}{1\linewidth} 
		\label{tab:fast_bench}
		\centering
		\scriptsize
		        \begin{tabular}{| c | c  ccccc }
                    \toprule
                    \multicolumn{1}{c}{Method} & \multicolumn{1}{c}{MOTA$\uparrow$} & \multicolumn{1}{c}{HOTA$\uparrow$} & \multicolumn{1}{c}{IDF1$\uparrow$} & \multicolumn{1}{c}{FP$\downarrow$} & \multicolumn{1}{c}{FN$\downarrow$} & \multicolumn{1}{c}{IDs$\downarrow$} \\
                    \hline
                    \multicolumn{1}{c}{ByteTrack \cite{zhang2022bytetrack} } & 56.8 & 60.0 & 77.5 & 3656 & 68552 & 533
                    \\
                    \multicolumn{1}{c}{StrongSORT \cite{du2023strongsort}} & 61.1 & 58.0 & 77.0 & 33989 & \textbf{64882} & 422
                    \\
                    \multicolumn{1}{c}{FeatureSORT \cite{hashempoor2024featuresort}} & 61.4  & 59.7 & 77.4 & \textbf{26993} & 74692 & 390\\
                    \hline
                    \hline
                    \rowcolor{gray!20} 
                    \multicolumn{1}{c}{FastTracker} & \textbf{64.4}  & \textbf{61.5} & \textbf{79.2} & 29730 & 68541 & \textbf{251} \\
                    \hline
                \end{tabular}
	\end{minipage}
\end{table}

\section{Discussion}
\label{sec:discussion}
FastTracker operates in real time across all evaluated scenarios, sustaining frame-rate processing on the incoming streams even under severe occlusions and dense object/frame conditions. This efficiency enables deployment on a wide range of edge devices without sacrificing accuracy. 
While FastTracker demonstrates strong performance across public and internal benchmarks, some limitations remain. Currently, the system relies on manually defined ROI regions and cone direction constraints, which must be specified using exactly four edges. This configuration may limit deployment in complex or dynamic scenes where such annotations are impractical or insufficient. As a promising direction for future work, recent advances in semantic segmentation and scene understanding can be leveraged to enable automatic detection of road boundaries, crosswalks, and other contextual cues, removing the need for manual setup. Moreover, extending the system to support arbitrary polygonal ROIs or more flexible directional constraints would allow broader applicability in real-world environments such as intersections, roundabouts, and multi-lane roads.

\section{Conclusion}
\label{sec:conclusion}
We introduce FastTracker, a very fast and lightweight multi-object tracker that operates without any CNN-based re-identification network. It effectively handles occlusions and leverages environment-aware cues such as spatial constraints to improve tracking accuracy. While designed for online deployment, it also supports optional post-processing for further refinement. Despite its simplicity, FastTracker outperforms most state-of-the-art methods and is suitable for deployment on resource-constrained devices.

% {\appendix[Proof of the Zonklar Equations]
% Use $\backslash${\tt{appendix}} if you have a single appendix:
% Do not use $\backslash${\tt{section}} anymore after $\backslash${\tt{appendix}}, only $\backslash${\tt{section*}}.
% If you have multiple appendixes use $\backslash${\tt{appendices}} then use $\backslash${\tt{section}} to start each appendix.
% You must declare a $\backslash${\tt{section}} before using any $\backslash${\tt{subsection}} or using $\backslash${\tt{label}} ($\backslash${\tt{appendices}} by itself
%  starts a section numbered zero.)}

 \bibliographystyle{IEEEtran}  % appearance order
\bibliography{References.bib}

\newpage

 

% \end{IEEEbiography}


% \vfill

\end{document}